\documentclass[mnsc,nonblindrev]{informs3} 

\OneAndAHalfSpacedXI 
\usepackage{hyperref}
\usepackage{bbm}
\usepackage{xfakebold}

\usepackage{endnotes}
\let\footnote=\endnote

\usepackage{natbib}
 \bibpunct[, ]{(}{)}{,}{a}{}{,}%
 \def\bibfont{\small}%

\usepackage{algorithm}
\usepackage{algorithmic}
\usepackage{subfigure}
\usepackage{graphicx}
\usepackage{makecell}

\usepackage{ulem}
\usepackage{booktabs}
\usepackage{multirow}
\usepackage{comment}

\usepackage{hyperref}

\usepackage{xcolor}
\hypersetup{
    colorlinks,
    linkcolor={red!50!black},
    citecolor={blue!50!black},
    urlcolor={blue!80!black}
}
\TheoremsNumberedThrough     
\ECRepeatTheorems

\EquationsNumberedThrough    

\newenvironment{assumption'}[1]
  {\renewcommand{\theassumption}{\arabic{assumption}$'$}%
   \addtocounter{assumption}{-1}%
   \begin{assumption}}
  {\end{assumption}}

\newcommand{\fbseries}{\unskip\setBold\aftergroup\unsetBold\aftergroup\ignorespaces}
\usepackage{amsmath}
\usepackage{amsfonts}
\usepackage{amssymb}
\usepackage{dsfont}

\begin{document}




\TITLE{Large Language Model Enhanced Machine Learning Estimators for Classification}




\ARTICLEAUTHORS{%
\AUTHOR{Yuhang Wu\textsuperscript{1}, Yingfei Wang\textsuperscript{2}, Chu Wang\textsuperscript{3}, and Zeyu Zheng\textsuperscript{1}}
\AFF{\textsuperscript{1}Department~of Industrial Engineering and Operations Research,  University of California Berkeley \\
\textsuperscript{2}Foster School of Business, University of Washington\\
\textsuperscript{3}Amazon }
 }

\ABSTRACT{%
Pre-trained large language models (LLM) have emerged as a powerful tool for simulating various scenarios and generating output given specific instructions and multimodal input. In this work, we analyze the specific use of LLM to enhance a classical supervised machine learning method for classification problems. We propose a few approaches to integrate LLM into a classical machine learning estimator to further enhance the prediction performance. We examine the performance of the proposed approaches through both standard supervised learning binary classification tasks, and a transfer learning task where the test data observe distribution changes compared to the training data. Numerical experiments using four publicly available datasets are conducted and suggest that using LLM to enhance classical machine learning estimators can provide significant improvement on prediction performance. 
}%


%

\maketitle

%


\section{Introduction}

Classification is a fundamental task in supervised machine learning, common across a wide array of applications. It involves training a model on a dataset where each instance is assigned a specific class label. When presented with a new, unlabeled instance, the trained model is expected to accurately predict the instance's class.

To illustrate, consider a classification problem aimed at predicting whether a customer will find a product relevant online. In this problem, each instance involves two-fold information on (i) customer needs, e.g., reflected by the search queries of the customer and other attributes, and (ii) product attributes, including title, description, and/or images of the product. In the training dataset, each instance is labelled with one of the two classifying categories: ``relevant" or ``not relevant". 

There are a wide range of well-established machine learning methods for such classification tasks, ranging from logistic regression, tree-based methods, to neural-networks, among many other; see \cite{efron2021computer}. These methods are typically trained and calibrated using the training dataset. Once trained, they serve as estimators to predict the labels of new, unlabeled instances. 

On the other hand, the emergence of pre-trained large language models (LLMs) offers an additional approach to these classification tasks, capable of functioning as estimators with or without additional fine-tuning on task-specific data. For instance, GPT \cite{radford2018improving} by OpenAI can serve as an estimator by simply taking an instance (comprising customer query and product information) as input and generating a prediction on whether the product is relevant to the customer's needs. Moreover, prompt engineering and fine-tuning can further enhance the model's performance on specific classification tasks.

This work is motivated by the following questions. Can we integrate LLM to classical machine learning methods to significantly enhance the performance, compared to the separate use of LLM or the separate use of a classical machine learning model? What are the different ways of doing such integration, and how do they perform compared to benchmarks? 

We deliver the following results and analysis in this work. \begin{enumerate}
    \item We analyze the linear combination of an LLM model and a machine learning (ML) model. We observe that generally LLM predictions are more reliable than that of a machine learning model on borderline data. We then develop an adaptive weighted linear combination of LLM and ML to further enhance the performance via the heavier use of LLM on those regions where ML shows less confidence.
    \item By treating LLM predictions as additional group information, we apply model calibration methods to classical machine learning models. This method is straightforward to carry out and can be used upon any classical machine learning model. 
    \item We consider the integration of LLM and ML on a transfer learning classification tasks. For the transfer learning task with covariate shift, we improve the machine learning model by augmenting the training data with additional samples from the target distribution, where the labels of those samples are generated by LLM. We then train a machine learning model on the augmented dataset to obtain better performance on the target distribution.
    \item We illustrate the empirical performances of our methods on four public datasets, including tasks such as relevance prediction, emotion recognition and hate speech detection. Numerical results show that all our methods perform better than only using LLM or only using a classical machine learning model.
\end{enumerate}

We would like to add some discussions before proceeding to the main sections of this work. The advantage of leveraging LLMs to enhance classical ML methods comes from two primary sources. Firstly, LLMs can serve as a variance reduction tool in addition to a classical machine learning model trained on the dataset. The use of LLMs in this context draws a close analogy to the method of control variates as a variance reduction tool in simulation literature; refer to \cite{asmussen2007stochastic}.  We also note that the use of LLM in a classical data-driven method creates additional needs for input data uncertainty analysis, and leave that for future discussions; see \cite{song2014advanced,feng2019efficient}. Secondly, LLMs can enhance model accuracy by leveraging their knowledge on a broader range of data; see \cite{moller2023prompt,gao2023chat,chen2024exploring}. This aspect is particularly beneficial for improving classical ML models in transfer learning tasks.











\section{Problem formulation and notation}
In this work, we focus on binary classification problems, using the task of predicting relevance as a representative example to illustrate our approach. The mathematical formulation for this task is consistent with other tasks, and we will further discuss in the numerical experiments section. The classification task is formulated as follows. The training data are given by $\left\{(\text{query}_{i},\text{product}_{i},\text{label}_i)\right\}_{i=1}^n$. Here, $\text{query}_{i}$ consists of the searching content of a customer, e.g. ``modern outdoor furniture".  Next, $\text{product}_{i}$ consists of a set of product information that includes product description, product features and possibly product image. The $\text{label}_i$ denoted the true relevance label of ``relevant" or ``irrelevant", usually obtained by manual annotation validated by several independent human annotators. For classification tasks, embeddings are utilized to derive embedded vectors of $\text{query}_{i}$ and $\text{product}_{i}$, denoted by $v_{i1}\in\mathbb{R}^{d_1},v_{i2}\in\mathbb{R}^{d_2}$. The feature covariate $x_i$ is then given by concatenating $v_{i1}$ and $v_{i2}$ as $x_i=\mathrm{concat}(v_{i1},v_{i2})\in\mathbb{R}^d$, where $d=d_1+d_2$. In text-based classification, it's also common to concatenate text strings using a special token before vectorization. We do not consider the fine-tuning of LLM at this stage, and simply consider the integration of a simple machine learning model and a pre-trained LLM model without fine tuning. 

The training dataset is then given by $\mathcal{D}_\mathrm{train}=\left\{(x_i,y_i)\right\}_{i=1}^{n}$, where $x_i\,(1\leq i\leq n)\in \mathbb{R}^d$ are features and $y_i\,(1\leq i\leq n)\in \{0,1\}$ are the relevance label. Here it is set that $y_i=0$ if $\text{label}_i$ is relevant, and $y_i=1$ for the irrelevant label. This formulation is generic for a wide range of binary classification problems.

One can train a classical machine learning model $\hat{f}_n(\cdot)$ on $\mathcal{D}_\mathrm{train}$, where $\hat{f}_n:\mathbb{R}^d\to[0,1]$ maps the feature $x_i$ to a score between $0$ and $1$, where the score represents the chance of irrelevance. If $\hat{y}_i=\hat{f}_n(x_i)\leq 0.5$, one predicts ``relevant" and otherwise ``irrelevant". The test set is given by $\mathcal{D}_\mathrm{test}=\{(x_i,y_i)\}_{i=n+1}^{n+m}$ and a trained machine learning model is evaluated by ${L}_\mathrm{test}=\frac{1}{m}\sum_{i=n+1}^{n+m}l\left(y_i,\hat{y}_i\right)$ for some loss function $l(\cdot,\cdot)$. In addition to the classical  machine learning model, in this work we also use a pre-trained large language model (LLM) to predict $y_i$. The LLM is given the input of a query-product pair and asked to output a score between 0 and 1 representing how likely the LLM evaluates the query and product as irrelevant. We denote the output of LLM for the $i$-th pair as $z_i\in[0,1]$ for $i=1,\cdots, n+m$.

\section{Integrating LLM into a Classical ML Estimator}\label{sec:estimator}

One naive way to utilize LLM is to treat it as a trained model, so $z_i\,(1\leq i\leq n+m)$ are just estimators of $y_i\,(1\leq i\leq n+m)$, in addition to the classical trained machine learning (ML) model, e.g., logistic regression. Then a first thought is to combine the two models (LLM and classical ML) to create a better ensemble model (noted as LLM-ML in this work). This is a standard combination task and can be executed by several methods. We illustrate such combination task by examining a simple linear combination of ML and LLM estimators. We first discuss a straightforward approach using a fixed constant weight. We then demonstrate through empirical data analysis that the performance can be enhanced by employing an adaptive weighting strategy, which adjusts the weights according to the prediction scores generated by the ML estimator. Such adaptive weighting strategy carries the thought that LLM may be more heavily used for instances where the classical ML is not that certain of. 
\subsection{Linear combination of ML and LLM estimators}\label{subsec:naive_linear}
 To combine the ML estimator and LLM estimator, one simple method is to consider the linear combination with constant weight, i.e.,
\begin{equation}
    \hat{y}^\mathrm{Linear}_i = \alpha \hat{y}_i+(1-\alpha)z_i,
\end{equation}
with some weight $\alpha\in [0,1]$.

The choice of $\alpha$ can be determined through standard cross validation procedure as follows. By randomly splitting $\mathcal{D}_\mathrm{train}$ as $\cup_{i=j}^k\mathcal{D}_\mathrm{train}^{j}$ for some $k\in\mathbb{N}_+$, we can train $k$ machine learning models $\hat{f}^j_n(\cdot)$ for $j=1,\cdots,k$, where $\hat{f}^j_n(\cdot)$ is trained on $\cup_{i=1,i\neq j}^k\mathcal{D}_\mathrm{train}^{i}$. Then the best weight $\alpha^*$ is given by
\begin{equation}\label{eq:best_const_alpha}
    \hat{\alpha}_n= \arg\min_{\alpha\in[0,1]} \frac{1}{n}\sum_{j=1}^k\sum_{(x_i,y_i)\in \mathcal{D}_\mathrm{train}^{j}} l\left(\alpha\hat{f}^j_n(x_i)+(1-\alpha)z_i,y_i\right).
\end{equation}
If we omit the superscript $j$ and denote $\hat{f}^j_n(x_i)$ by $\hat{y}_i^\mathrm{cv}$, for the $l_2$ loss $l(x,y)=(x-y)^2$, this procedure is equivalent to regress $\{y_i-z_i\}_{i=1}^n$ on $\{\hat{y}_i^\mathrm{cv}-z_i\}_{i=1}^n$, which gives $\hat{\alpha}_n^{l_2}=\frac{\sum_{i=1}^n(y_i-z_i)(\hat{y}_i^\mathrm{cv}-z_i)}{(\hat{y}_i^\mathrm{cv}-z_i)^2}$. Informally speaking, if we assume $(x_i,z_i,y_i)\stackrel{\mathrm{i.i.d.}}\sim P$ for some distribution $P$ on $\mathbb{R}^d\times [0,1]^2$, and assume that $\hat{f}_n$ converges to some $f_0$ as $n\to +\infty$, then we will have 
$$\hat{\alpha}_{n,l_2}\to \alpha^*=\frac{\mathbb{E}_{P}\left[(Y-Z)(f_0(X)-Z)\right]}{\mathbb{E}_{P}\left(f_0(X)-Z\right)^2},\text{ as }n\to+\infty,$$
which is the minimizer of the mean-squared error
   $\mathrm{MSE}_\mathrm{L}(\alpha)= \mathbb{E}\left[\alpha f_0(X)+(1-\alpha)Z-Y\right]^2,$
so $\mathrm{MSE}_\mathrm{L}(\alpha^*)\leq \mathrm{MSE}_\mathrm{L}(0)=\mathbb{E}(Z-Y)^2$ and $\mathrm{MSE}_\mathrm{L}(\alpha^*)\leq \mathrm{MSE}_\mathrm{L}(1)=\mathbb{E}(f_0(X)-Y)^2$, which means the asymptotic behaviour of the LLM-ML ensemble estimator is no worse than two baseline estimators.

\subsection{Linear combination with adaptive weights}\label{subsec:adaptive_linear}
\begin{figure}
    \centering
    \includegraphics[scale =0.16]{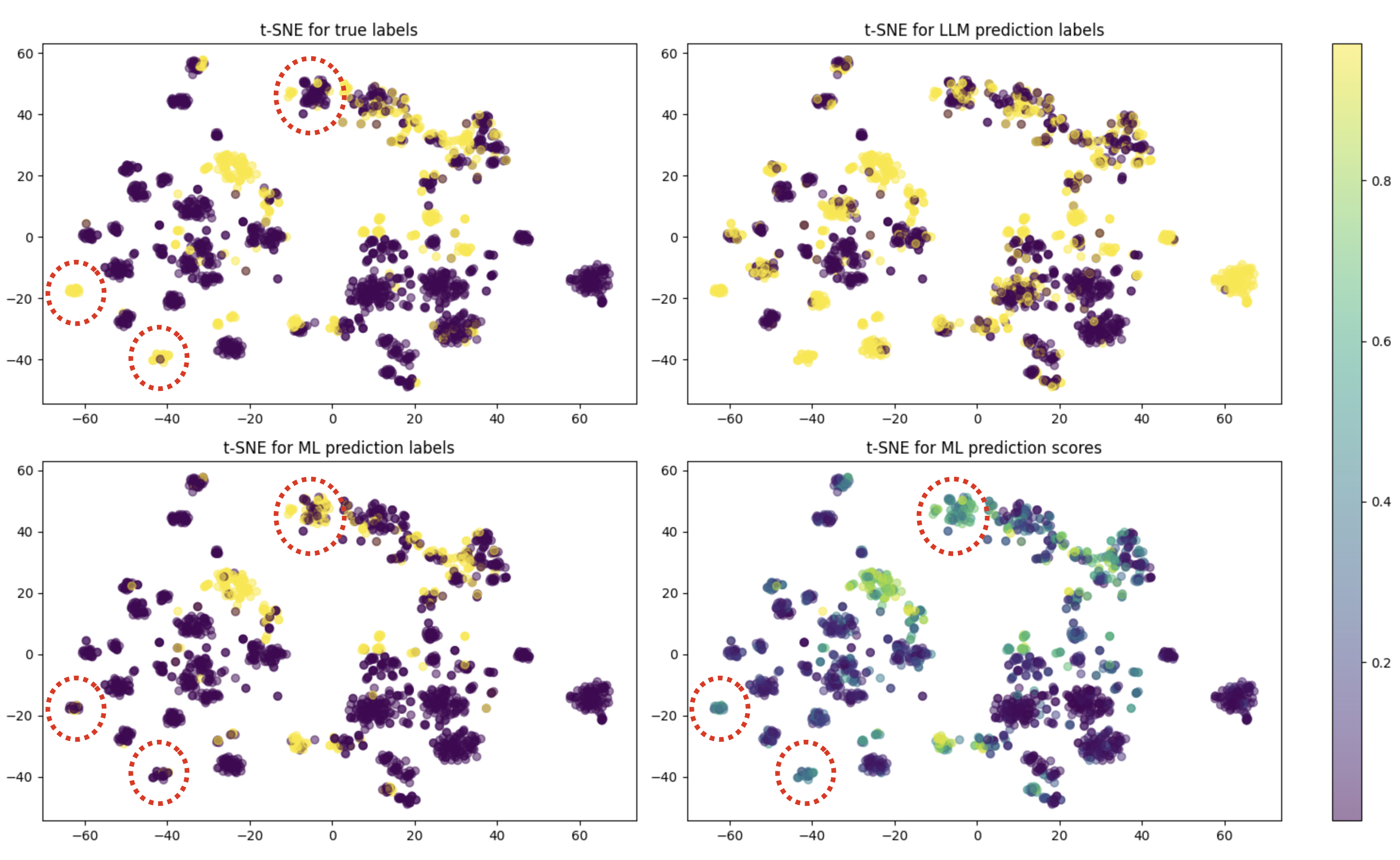}
    \caption{t-SNE visualization of an illustration data set}
    \label{fig:tsne}
\end{figure}

One can improve this method by taking $\alpha$ to be dependent on $\hat{y}_i$ as $\alpha(\hat{y}_i)$. Consider the following visualization results in Figure~\ref{fig:tsne}. Here t-SNE \cite{hinton2002stochastic} is a statistical method for visualizing high-dimensional data by giving each sample a location in a two-dimensional map. We can see the classical machine learning model sometimes does not do well on the boundary of relevant samples and irrelevant samples, and many of such errors are made when the prediction score $\hat{y}_i$ is close to $0.5$. On the contrary, the LLM model shows different patterns, in the sense that its accuracy is relatively stable even for those borderline data. We refer to the red circles in the figure for some illustration. This motivates us to take $\alpha(\hat{y}_i)$ to be larger when $\hat{y}_i$ is close to $0$ or $1$, and if $\hat{y}_i$ is close to $0.5$, which indicates $x_i$ is likely to located on the boundary, we set it to be smaller. 

In this case, the LLM-ML estimator is given as an adaptive weighted linear combination of two estimator:
\begin{equation}
\hat{y}^\mathrm{AL}_i = \alpha(\hat{y}_i)\cdot \hat{y}_i+\left(1-\alpha(\hat{y}_i)\right)\cdot z_i.
\end{equation}
We would expect $\alpha(\cdot):[0,1]\to[0,1]$ to be a decreasing function on $[0,0.5]$, and then increases on $[0.5,1]$. By taking a predetermined hypothesis space $\mathcal{A}\subset \{\alpha\vert\alpha:[0,1]\to [0,1]\}$ for all possible choices of $\alpha(\cdot)$ we are interested, the choice of $\alpha(\cdot)$ can similarly be determined through cross validation as what we just did for a fixed $\alpha$. To see this, we can now modify \eqref{eq:best_const_alpha} as:

\begin{equation}\label{eq:best_func_alpha}
    \hat{\alpha}_n= \arg\min_{\alpha\in\mathcal{A}} \frac{1}{n}\sum_{i=1}^n l\left(\alpha\left(\hat{y}_i^\mathrm{cv}\right)\cdot\hat{y}_i^\mathrm{cv}+\left(1-\alpha\left(\hat{y}_i^\mathrm{cv}\right)\right)\cdot z_i,y_i\right),
\end{equation}
where $\hat{y}_i^\mathrm{cv}=\hat{f}^j_n(x_i)$ and by slightly abusing the notation $\hat{\alpha}_n$ is now a function in $\mathcal{A}$. 

We now consider the choice of $\mathcal{A}$. A large $\mathcal{A}$ such as all continuous functions from $[0,1]$ to $[0,1]$ $\{\alpha\vert \alpha:[0,1]\to[0,1],\alpha\in C\}$ will make the optimization problem in \eqref{eq:best_func_alpha} intractable and can easily lead to overfit, but a simple $\mathcal{A}$ such as all constant functions on $[0,1]$ will just give a fixed weight and cannot fully use the strength of LLM. In this work, we consider the class of piecewise constant functions, i.e.
\begin{equation}
    \mathcal{A}_r=\{\alpha\vert\alpha(x)=\sum_{i=1}^r w_i\mathbf{1}_{\{x\in I_i\}}\text{ with }w_i\in[0,1](i=1,\cdots,r)\},
\end{equation}
here $r\in\mathbb{N}_+$ is a pre-determined positive integer and $I_i(1\leq i\leq r)$ is a partition of $[0,1]$. For simplicity, we take $I_i=[\frac{i-1}{r},\frac{i}{r})$ for $1\leq i\leq r-1$ and $I_r=[\frac{r-1}{r},1]$. In this case, \eqref{eq:best_func_alpha} can now be formulated as:
\begin{equation}
\begin{aligned}
        \hat{\alpha}_{n,r}&= \arg\min_{\alpha\in\mathcal{A}_r}\frac{1}{n}\sum_{i=1}^n l\left(\alpha\left(\hat{y}_i^\mathrm{cv}\right)\cdot\hat{y}_i^\mathrm{cv}+\left(1-\alpha\left(\hat{y}_i^\mathrm{cv}\right)\right)\cdot z_i,y_i\right)\\
        &=\arg\min_{w_1,\cdots,w_r}\frac{1}{n}\sum_{j=1}^r\sum_{i:\hat{y}_i^\mathrm{cv}\in I_j}l\left(w_j\hat{y}_i^\mathrm{cv}+\left(1-w_j\right) z_i,y_i\right).\\
\end{aligned}
\end{equation}
This is equivalent to solve $r$ optimization problems for samples with indices in $\{i\vert \hat{y}_i^\mathrm{cv}\in I_j\}(j=1,\cdots,r)$ separately, and each of them is what we just discussed in last part. If we are using $l_2$ loss then clearly this piecewise constant weight function gives an MSE no larger than any fixed constant weight.      

\begin{algorithm}[htbp]
\caption{Adaptive weighted linear combination of LLM and ML}\label{alg:AdaLinear}
\begin{algorithmic}[1]
    \STATE \textbf{Input: } $\mathcal{D}_\mathrm{train}=\{(x_i,z_i,y_i)\}_{i=1}^n$, Number of pieces $r$, and a partition of $[0,1]$ $I_i(1\leq i\leq r)$;
    \STATE Train a ML model $f(\cdot)$ on $\{(x_i,y_i)\}_{i=1}^n$ and construct cross validation predictions $\{(x_i,\hat{y}_i^\mathrm{cv})\}_{i=1}^n$.
    \STATE Solve subproblems $w_j=\arg\min_{w}\sum_{i:\hat{y}_i^\mathrm{cv}\in I_j}l\left(w\hat{y}_i^\mathrm{cv}+\left(1-w\right) z_i,y_i\right)$ for $j=1,\cdots,r$.
    \STATE \textbf{Output}: the weight function $\alpha(x)=\sum_{i=j}^r w_j\mathbf{1}_{\{x\in I_j\}}$.
\end{algorithmic}
\end{algorithm}

\section{Calibration with LLM}\label{sec:calibration}
In this section, we consider calibrating a classical ML model (e.g., logistic regression) with LLM. We start by introducing some backgrounds on calibration, then we discuss how to use LLM to calibrate a classical ML model.

\subsection{Background on calibration}
Suppose our data are realizations of $(X,Y)\sim\mathcal{D}$ with $Y\in\{0,1\}$, and the model is $f$. Ideally, we hope to find a model such that
\begin{equation}
    f(x)=P(Y=1\vert X=x). \text{ (Informally)}
\end{equation}
However, this is generally unrealistic and there are at least two reasons. First, for each fixed $x$, the label is already determined, and there is no randomness so the probability in the RHS is not well-defined. Second, the domain of $X$ in our scenario consists of pairs of query and item, and each of them comes from an extremely large set of English words, so it is generally impossible to find such a function with limited samples. Calibration is a  tractable condition that can be viewed as a coarsening of the condition above.

\begin{definition}\label{def:calibration}
    (Calibration) For $(X,Y)\sim\mathcal{D}$, the bias of $f$ at the $p$-th level set is defined as
    \begin{equation}
        \Delta_p(f)\triangleq\mathbb{E}_\mathcal{D}[Y-f(X)\vert f(X)=p].
    \end{equation}
     If $\Delta_p(f)\equiv0$, we say that $f$ is calibrated w.r.t. $\mathcal{D}$. We omit the subscript $\mathcal{D}$ when there is no ambiguity.
\end{definition}

Intuitively, calibration can be considered as a minimal condition for a model $f$ to be good, in the sense that the probability of $Y=1$ conditional on $f(x)=p$ is indeed $p$. For each model $f$, if it is not calibrated, we can define $\hat{f}(x)=f(x)+\Delta_{f(x)}(f)$ and it is straightforward to verify that $\hat{f}$ is calibrated. However, it is infeasible to conditional on $f(X)=p$ in practice with finite samples, and a simple strategy is to do discretization on a uniform grid and calibrate $f$ on the grid. Specifically, we define the uniform grid $[\frac{1}{M}]=\{\frac{i}{m}\}_{i=0}^M$ for some $M\in\mathbb{N}_+$, and define
\begin{equation}\label{eq:grid_calibration}
    f'(x)=\arg\min_{u\in[\frac{1}{M}]}\vert f(x)-u\vert.
\end{equation}
This rounds the output of $f$ to the grid $[\frac{1}{M}]$ and can be viewed as a discretization of $[0,1]$. Now, for our data $\{(x_i,y_i)\}_{i=1}^n$, we can construct the calibration of $f$ as
\begin{equation}
    \hat{f}^{(M)}(x) = f(x)+\hat\Delta_{f'(x)}(f), 
\end{equation}
where $\hat\Delta_{f'(x)}(f)$ can be estimated through data by calculating the mean of $y_i-f(x_i)$ conditional on the value of $f'(x_i)$. We refer to \cite{roth2022uncertain} for more theoretical results of $\hat{f}^{(M)}$.

\subsection{Calibrate ML model with LLM}\label{subsec:calibrate}
Above calibration procedure only involves the model $f$, and we now integrate LLM into this procedure. In this scenario, we assume our data are realizations of $(X,Z,Y)\sim P$, where $Z\in[0,1]$ is the output of LLM. We have a ML model $f$ and we want to use it to construct a improved model, which takes $X$ and $Z$ as inputs and generates a prediction for $Y$. We start by introducing an enhanced version of Definition \ref{def:calibration}, which is also known as the ``multi-accuracy" condition in \cite{kim2019multiaccuracy}.
\begin{definition}\label{def:multiaccuracy}
    (Multi-accuracy with LLM) Suppose $(X,Z,Y)\sim P$, A model $f$ is multi-accurate w.r.t. $Z$ if
    \begin{equation}\label{eq:multi_accuracy}
        \mathbb{E}_P\left[Y-f(X)\vert Z\right]= 0.
    \end{equation}
\end{definition}
If we treat $Z$ as a covariate, then Definition \ref{def:multiaccuracy} can be viewed as a condition for a model $f$ to be consistent conditional on the covariate. Now, with both Definition \ref{def:calibration} and \ref{def:multiaccuracy}, we want to construct some $\hat{f}^{M,M'}$ from $f$ such that $\hat{f}^{M,M'}$ is calibrated and multi-accurate. Again it is infeasible to conditional on $Z$ if $Z$ is continuously distributed with finite samples, so we consider a discretization similar as what we did in \eqref{eq:grid_calibration}. Let $[\frac{1}{M'}]=\{\frac{i}{m}\}_{i=0}^M$ for some $M'\in\mathbb{N}_+$, and define
\begin{equation}
    S_{p,q}(f) = \left\{(x,z)\Big\vert \arg\min_{u\in[\frac{1}{M}]}\vert u-f(x)\vert=p,\arg\min_{v\in[\frac{1}{M'}]}\vert v-z\vert =q\right\}
\end{equation}
for $p=\frac{i}{M}(0\leq i\leq M)$ and $q=\frac{j}{M'}(0\leq j\leq M')$. Also define
\begin{equation}
\Delta_{p,q}(\hat{f},f)\triangleq \mathbb{E}\left[Y-\hat{f}(X)\vert S_{p,q}(f)\right]. 
\end{equation}
Then one simple way to construct some $\hat{f}^{M,M'}$ is to make sure $\Delta_{p,q}(\hat{f}^{M,M'},f)=0$ for all $p,q$. Informally speaking, if $\Delta_{p,q}(\hat{f}^{M,M'},f)=0$ for all $p,q$ then $\hat{f}^{M,M'}$ is calibrated and multi-accurate up to some discretization error. This naturally leads to the following choice:
\begin{equation}\label{eq:naive_multicalibration}
    \hat{f}^{M,M'}_1(x)=f(x)+\hat\Delta_{p,q}(f,f),  \text{ if }(x,z)\in S_{p,q}(f), 
\end{equation}
where
\begin{equation}
    \hat{\Delta}_{p,q}(f,f)=\frac{\sum_{1\leq i\leq n,(x_i,z_i)\in S_{p,q}(f)}\left(y_i-f(x_i)\right)}{\sum_{1\leq i\leq n,
    (x_i,z_i)\in S_{p,q}(f)} 1}
\end{equation}
is the empirical mean of $Y-f(X)$ conditional on $S_{p,q}(f)$ with samples in the training set. Then, if we assume $\{(x_i,z_i,y_i)\}_{i=1}^{n+m}\stackrel{\text{i.i.d.}}\sim P$, we have
\begin{equation}
\begin{aligned}
    &\Delta_{p,q}(\hat{f}^{M,M'}_1,f)\\
    =&\mathbb{E}\left[Y-\hat{f}^{M,M'}(X)\big\vert S_{p,q}(f)\right]\\
    =&\mathbb{E}\left[Y-f(X)\vert S_{p,q}(f)\right]-\mathbb{E}\left[\hat\Delta_{p,q}(f,f)\vert S_{p,q}(f)\right]\\
    =&0,
\end{aligned}
\end{equation}
which is exactly what we want.

While \eqref{eq:naive_multicalibration} gives one way to construct a function that is both calibrated and multi-accurate up to some discretization error, the problem is that it involves calculating $(M+1)\times (M'+1)$ conditional means for $S_{p,q}(f)$. When $M$ and $M'$ are large, this quantity may be of the same order of the sample size $n$ or even larger, making the estimation of conditional means not precise enough. This motivates the following choice. We take
\begin{equation}\label{eq:group_multicalibration}
    \hat{f}^{M,M'}_2(x)=f(x)+\hat{w}_{i,0}+\hat{w}_{j,1}, \text{ if }(x,z)\in S_{\frac{i}{M},\frac{j}{M'}}(f), 
\end{equation}
where $\hat{w}_{i,0}(0\leq i\leq M)$ and $\hat{w}_{j,1}(0\leq j\leq M')$ are determined by the following optimization problem:
\begin{equation}
    \hat{w}_{0,0},\cdots,\hat{w}_{M,0},\hat{w}_{0,1},\cdots,\hat{w}_{M',1}=\arg\min_{\substack{w_{0,0},\cdots,w_{M,0},\\ w_{0,1},\cdots,w_{M',1}}}\sum_{i=1}^n\left(\hat{f}^{M,M'}_2(x_i)-y_i\right)^2.
\end{equation}
Since this is a least square linear regression problem, the solution always exists. For any solution, the optimality conditions give:
\begin{equation}
        \sum_{f'(x_i)=\frac{j}{M}}\left(\hat{f}^{M,M'}_2(x_i)-y_i\right)=0, \quad \forall~0\leq j\leq M,
\end{equation}
so $\mathbb{E}\left[Y-\hat{f}^{M,M'}_2(X)\Big\vert f'(X)\right]=0$, which implies $\hat{f}^{M,M'}_2$ is calibrated up to some discretization error. Similarly, it is also multi-accurate up to some discretization error. The construction of $\hat{f}^{M,M'}_2$ only requires the estimation of $M+M'+2$ parameters, which is much smaller than $(M+1)\times (M'+1)$ parameters for $\hat{f}^{M,M'}_1$, and can thus mitigate issues such as large estimation error and overfitting in $\hat{f}^{M,M'}_1$. 

\begin{algorithm}[htbp]
\caption{Calibrate ML model with LLM}\label{alg:calibration}
\begin{algorithmic}[1]
    \STATE \textbf{Input: } $\mathcal{D}_\mathrm{train}=\{(x_i,z_i,y_i)\}_{i=1}^n$, $M,M'\in \mathbb{N}_+$;
    \STATE Train a ML model $f(\cdot)$ on $\{(x_i,y_i)\}_{i=1}^n$.
    \STATE \textbf{Output}: the calibration of $f(\cdot)$ given by $\hat{f}_1^{M,M'}(\cdot)$ in \eqref{eq:naive_multicalibration} or $\hat{f}_2^{M,M'}(\cdot)$ in \eqref{eq:group_multicalibration}.
\end{algorithmic}
\end{algorithm}

\section{Transfer learning with LLM}\label{sec:transfer}
Recall that in Figure~\ref{fig:tsne} we saw that LLM can perform more stable than a classical ML model, this motivates us to apply LLM to transfer learning, where the training distribution of the model can be different from that of its application. Such transfer learning tasks can be challenging to a classical ML model.

Again consider the relevance label prediction task, suppose that there are no or very few samples in the labeled training set that contains products related to ``bed", but the test datasets contain a large number of products related to ``bed". This can happen in applications where a new type of product is introduced to the platform. We would expect that the classical ML model may not do well on the test datasets because the model has not seen much information about ``bed". One possible remedy is to utilize LLM to augment the training set for the bed category by adding data labeled with predictions from the LLM. This can be formulated as follows. Suppose we have a training set $\mathcal{D}_\mathrm{train}=\{(x_i,y_i)\}_{i=1}^n\stackrel{\text{i.i.d.}}\sim P_1$, and we train a ML model $\hat{f}_n\in\mathcal{F}$ on $\mathcal{D}_\mathrm{train}$ for some model class $\mathcal{F}$. We want to evaluate its performance by its expected loss on $P_2$ w.r.t. some loss function $l$: $\mathbb{E}_{(X,Y)\sim P_2}\left[l(\hat{f}_n(X),Y)\right]$. Here we assume $P_1=P_{X,1}\times P_{Y\vert X}$ and $P_2=P_{X,2}\times P_{Y\vert X}$, where $P_{X,l}~(l=1,2)$ are two distributions of $X$ on its domain $\mathcal{X}$ and the conditional distribution of $Y\vert X$ is given by $P_{Y\vert X}$ and is the same for $P_1$ and $P_2$. This scenario is usually referred as covariate shift \cite{sugiyama2007covariate}. While we do not have other labeled training data, we are accessible to sample more $X$ as we want, that is to say, we can augment the training set by adding $\{(x_i,z_i)\}_{i=n+1}^{n+m}$ for some $x_i$ and $m\in\mathbb{N}_+$ as we want. With the augmented training set, we can train a new ML model $\hat{f}_{n,m}$ as follows:
\begin{equation}
    \hat{f}_{n,m}=\arg\min_{f\in \mathcal{F}} \frac{1}{n+m}\sum_{i=1}^nl\left(f(x_i),y_i\right)+\frac{1}{n+m}\sum_{i=n+1}^{n+m}l_0(f(x_i),z_i).
\end{equation}
Here $l(\cdot,\cdot)$ is the loss we are interested, and $l_0(\cdot,\cdot)$ is some weak-supervised loss that can be viewed as a relaxation of $l(\cdot,\cdot)$. The rational is that the label of $\{x_i\}_{i=n+1}^{n+m}$ are generated from LLM and may be biased and noisy, so a relaxed loss such as $l_0(x,y)=\min_{\varepsilon\in[-a,a]}l\left(x,y+\varepsilon\right)$ for some $a\geq 0$ gives looser penalization on those samples.

Regarding the choice of $\{x_i\}_{i=n+1}^{n+m}$, suppose we sample them from some distribution $P_{X,3}$. Let the probability density function of $P_{X,3}$ be $p_i(x)$ for $i=1,2,3$, then in the ideal case we would expect 
\begin{equation}
    \frac{n}{n+m}p_1(x)+\frac{m}{n+m}p_3(x)\equiv p_2(x),
\end{equation} 
which then gives a choice of $P_3$ as
\begin{equation}\label{eq:p3}
    p_3(x)=p_2(x)+\frac{n}{m}\left(p_2(x)-p_1(x)\right).
\end{equation}
However, in order to make sure $p_3(x)\geq 0$, we need $m\geq \sup_{x}\frac{p_1(x)-p_2(x)}{p_2(x)}n$. When the quantity in the RHS is large, we will need lots of labels generated from LLM, which may not be a reasonable choice. In the extreme scenario that $p_2(x)=0$ and $p_1(x)>0$, it is impossible to find a $p_3(x)\in[0,1]$ satisfying \eqref{eq:p3}. Thus, in these scenarios, a heuristic choice such as $p_3(x)\propto \max\left(p_2(x)+\frac{n}{m}\left(p_2(x)-p_1(x)\right),0\right)$ or simply $p_3(x)=p_2(x)$ may be better. In Section \ref{subsec:transfer}, we illustrate our method in the scenario that the supports of $p_1(\cdot)$ and $p_2(\cdot)$ are disjoint, and we show that a naive choice of $p_3(x)=p_2(x)$ can already significantly improve the performance of the classical machine learning model.

\begin{algorithm}[htbp]
\caption{Transfer learning with LLM}\label{alg:transfer}
\begin{algorithmic}[1]
    \STATE \textbf{Input: } $\mathcal{D}_\mathrm{train}=\{(x_i,z_i,y_i)\}_{i=1}^n$ and a training distribution $P_1$, a target distribution $P_2$, $m\in\mathbb{N}_+$;
    \STATE Choose a sampling distribution $P_{X,3}$ based on $P_1$ and $P_2$, sample $\{x_i\}_{i=n+1}^{n+m}$.
    \STATE Label $\{x_i\}_{i=n+1}^{n+m}$ with LLM to obtain $\{(x_i,z_i)\}_{i=n+1}^{n+m}$
    \STATE Train a ML model $\hat{f}_{n,m}$ on $\{(x_i,y_i)\}_{i=1}^n\cup\{(x_i,z_i)\}_{i=n+1}^{n+m}$.
    \STATE \textbf{Output}: $\hat{f}_{n,m}$.
\end{algorithmic}
\end{algorithm}

\section{Numerical experiments}
We now illustrate the performances of the proposed LLM-ML integration methods through numerical experiments with public datasets. In Section \ref{subsec:dataset}, we provide the information of the datasets in use. In Section \ref{subsec:predict}, we compare the performances of the proposed methods in Section \ref{sec:estimator} and \ref{sec:calibration}. In Section \ref{subsec:transfer}, we consider the transfer learning task in Section \ref{sec:transfer}. Our code is available at \url{https://github.com/wyhArturia/llm_enhanced_ml}.
\subsection{Datasets}\label{subsec:dataset}
We illustrate our methods on four public datasets. For predicting relevance labels, we use the Wayfair Annotation DataSet
(WANDS) \cite{chen2022wands}. WANDS is a discriminative, reusable, and fair human-labeled dataset. It is one of the biggest publicly available search relevance dataset and is effective in evaluating and discriminating between different models. It consists of $480$ queries, $42994$ products, and $233000$ annotated query-product relevance labels. We also use the following three public datasets: Yelp's dataset (\url{https://www.yelp.com/dataset}), Emotion dataset \cite{saravia-etal-2018-carer} and Hate speech dataset \cite{de2018hate}. All of them are NLP classification tasks and have similar formulations as the relevance label prediction task as described in the manuscript. 

\subsection{Combined estimators and calibration with LLM}\label{subsec:predict}
We first illustrate the LLM-ML combined estimators in Section \ref{sec:estimator} and the calibrated estimators in Section \ref{sec:calibration} on four datasets. We randomly split each dataset into a training set and a testing set, and we compare the performances of following methods on testing sets:
\begin{enumerate}
    \item The large language model method (\textbf{LLM}): directly use GPT-3.5-Turbo-Instruct to classify the input instance. The prompt for each task consists of step-by-step instructions such as key indicators, evaluation rules, examples and quality checks. All prompts are provided in our code.
    \item The machine learning method (\textbf{ML}): for each training set which consists of word vectors and labels, we train a logistic regression model and apply it to the testing set for predictions. We use this simple machine learning model for illustration purposes, as we are mainly focus on how to use LLM to improve the ML model instead of the performance of the ML model.
    \item Naive linear combination of LLM and ML methods (\textbf{Linear}): our method in Section \ref{subsec:naive_linear} that uses a linear combination of LLM and ML estimators with a fixed weight. The weight is estimated through least square method on the training set.
    \item Adaptive weighted linear combination (\textbf{AdaLinear}): our method in Section \ref{subsec:adaptive_linear} that uses a linear combination of LLM and ML estimators with a piecewise constant weight function. The weight function is estimated through least square method on the training set. The number of pieces $r$ is also tuned on the training set, but its value will not affect the performance too much as long as it is not too small ($r=1$) or too large ($r>20$).
    \item Calibrated estimator with LLM (\textbf{Calibration}): our method in Section \ref{subsec:calibrate} that calibrates the ML estimator with LLM predictions. Since our LLM predictions are binary, we take $M'=2$. The value of $M$ is tuned on the training set. We use the construction in \eqref{eq:naive_multicalibration} for simplicity.
\end{enumerate}
The results are given as follows:

\begin{table}[h]
\centering
\caption{Test accuracy of different methods on four datasets}
\label{tab:dataset_performance}
\begin{tabular}{c|c|c|c|c|c}
& \textbf{LLM} & \textbf{ML} & \textbf{Linear} & \textbf{AdaLinear}$(r)$ & \textbf{Calibration}$(M)$ \\ \hline
WANDS & $0.775$ & $0.803$ & $0.838$ & $\mathbf{0.846}(4)$ & $0.840(10)$ \\
Yelp & $0.724$ & $0.691$ & $0.739$ & $0.742(4)$ & $\mathbf{0.743}(20)$ \\
Emotion & $0.759$ & $0.799$ & $0.806$ & $\mathbf{0.812}(10)$ & $0.806(10)$ \\
Hate & $0.672$ & $0.717$ & $0.720$ & $0.724(4)$ & $\mathbf{0.731}(12)$ \\
\end{tabular}
\end{table}
From the above results we can see that, on all four datasets, our three methods (Linear, AdaLinear and Calibration) that utilizing both ML and LLM estimators perform better than only use LLM or ML. The AdaLinear method performs better than the Linear method, which is reasonable as Linear is a special case of AdaLinear. AdaLinear and Calibration have comparable performances, indicating that both model ensemble and calibration are reasonable ways to enhance ML models with LLM.

\subsection{Transfer learning with LLM}\label{subsec:transfer}
We now consider the task of transfer learning with LLM in Section \ref{sec:transfer}. In order to illustrate the performance of our method, we manually construct a table dataset and a bed dataset from the WANDS dataset and are provided in our code. We compare the performances of following methods on testing sets:
\begin{enumerate}
    \item The large language model method (\textbf{LLM}): directly use GPT-3.5-Turbo to classify the input instances, similar as that in Section \ref{subsec:predict}.
    \item The machine learning model for the table dataset (\textbf{MLTable}): we train a logistic regression model on $1000$ word vectors in the table dataset. We would expect this ML model works well on the table dataset, but may not do well on the bed dataset.
    \item Naive linear combination of LLM and ML methods (\textbf{Linear}): our method in Section \ref{subsec:naive_linear} that uses a linear combination of LLM and MLTable methods with a fixed weight, similar as that in Section \ref{subsec:predict}. We use this as one baseline method. The reason we do not use an adaptive weight function is that, we can only calculate the weight function on the training set, which can lead to overfitting as the target distribution is different from the training set.
    \item The transfer learning method with LLM discussed in Section \ref{sec:transfer} (\textbf{Transfer}$(m)$): we randomly add $m$ bed samples and corresponding labels generated by LLM into our table dataset, then train a logistic regression model on all $10000+m$ word vectors. With the notations in Section \ref{sec:transfer}, this is equivalent to take $p_3(x)=p_2(x)$.
\end{enumerate}
We uses a table testing set and a bed testing set to evaluate these methods. The results are given as follows:
\begin{table}[h]
\centering
\caption{Test accuracy of four methods on two datasets}
\label{tab:transfer_results}
\begin{tabular}{c|c|c|c|c|c|c}
& \textbf{LLM} & \textbf{MLTable} & \textbf{Linear} & \textbf{Transfer}$(5000)$ & \textbf{Transfer}$(10000)$ &\textbf{Transfer}$(15000)$ \\ \hline
Table & $0.738$ & $0.915$ &  $\mathbf{0.923}$ & $0.910$ &$0.910$ & ${0.911}$\\
Bed & $0.753$ & $0.721$ & $0.737$ & $0.784$ & $\mathbf{0.786}$& $\mathbf{0.786}$ \\
\end{tabular}
\end{table}

We can see that the Transfer method always performs better than LLM and MLTable on the bed dataset with a small sacrifice on its performance on the table dataset, regardless of the choice of $m$. However, the choice of $m$ is subtle and can still affect the performance of our method. The linear combination of LLM and MLTable can perform well on the table dataset, but is not robust to the covariate shift and is not as good as our Transfer method on the bed dataset.

\section{Conclusion}
We conclude the work by discussing some unexplored aspects of this work. Our goal is to show that LLM can enhance a classification machine learning method for classification problems. We have selected the standard logistic regression method as the benchmark machine learning method. We believe that the exact size of improvement brought by LLM will vary by choosing different machine learning methods as benchmark. Also, the performance of LLM can vary by using different prompts. We did not push on prompt engineering to figure out the best way to design prompts to max out the LLM potential. The proposed approaches in this work may be further expanded by integrating prompt engineering and fine tuning with certain training data.

\bibliographystyle{informs2014}

\bibliography{bib.bib}

\end{document}